\pdfoutput=1
\documentclass[11pt]{article}

\usepackage[final]{naacl2021}

\usepackage{times}
\usepackage{latexsym}

\usepackage{graphicx}
\usepackage{grffile}
\usepackage{multirow}
\usepackage{xcolor,colortbl}
\usepackage{amsmath}
\usepackage{amssymb}
\usepackage{makecell}
\usepackage[super]{nth}
\usepackage{arydshln}

\usepackage{algorithm}
\usepackage[noend]{algpseudocode}
\usepackage{regexpatch}
\usepackage{subcaption}

\usepackage{microtype}

\usepackage[draft]{todo}
\newcommand{\fyTodo}[1]{\Todo[FY:]{\textcolor{orange}{#1}}}

\newcommand{\fyDone}[1]{\done[FY]\Todo[FY:]{\textcolor{orange}{#1}}}
\newcommand{\fyFuture}[1]{\done[FY]\Todo[FY:]{\textcolor{red}{#1}}}

\title{Code-Switching Translation}
\title{Disentangling Code-Switched data with Multi-Target Neural Machine Translation}
\title{Decoding Code-Switched data with Multi-Target Neural Machine Translation}
\title{Can You This \"Ubersetzen? Machine Translation for Code-Switched Input}
\title{Can You Traducir This? Machine Translation for Code-Switched Input}

\author{Jitao Xu  \\
    Univ. Paris-Saclay,   \\
    \& CNRS, LISN \\
    Orsay, France \\
  \texttt{jitao.xu@limsi.fr} \\\And
    Fran\c{c}ois Yvon  \\
    Univ. Paris-Saclay,   \\
    \& CNRS, LISN \\
    Orsay, France \\
  \texttt{francois.yvon@limsi.fr} \\}
\date{}

\begin{document}
\maketitle

\begin{abstract}
Code-Switching (CSW) is a common phenomenon that occurs in multilingual geographic or social contexts, which raises challenging problems for natural language processing tools. We focus here on Machine Translation (MT) of CSW texts, where we aim to simultaneously disentangle and translate the two mixed languages. Due to the lack of actual translated CSW data, we generate artificial training data from regular parallel texts. Experiments show this training strategy yields MT systems that surpass multilingual systems for code-switched texts. These results are confirmed in an alternative task aimed at providing contextual translations for a L2 writing assistant.
%Our analysis show that our systems implicitly embed automatic language identification (LID) abilities.
\fyDone{Write the abstract}
\end{abstract}

\section{Introduction\label{sec:introduction}}

Code-Switching (CSW) denotes the alternation of two languages within a single utterance \citep{Poplack80typology, Sitaram19survey}. It is a common communicative phenomenon that occurs in multilingual communities during spoken and written interactions. CSW is a well studied phenomenon in linguistic circles and has given rise to a number of theories regarding the structure of mixed language fragments \citep{Poplack1978syntactic, Pfaff1979constraints, Poplack80typology, Belazi1994code,Myers97duelling}. The Matrix Language Frame (MLF) theory \citep{Myers97duelling} defines the concept of \textit{matrix} and \textit{embedded} languages where the \textit{matrix language} is the main language that the sentence structure should conform to and notably provides the syntactic morphemes, while the influence of the \textit{embedded language}
% \footnote{CS may occur between more than two languages in a single utterance, a scenario that we do not consider here.}
is lesser and is mostly manifested in the insertion of content morphemes.

The rise of social media and user-generated content has made written instances of code-switched language more visible. It is estimated that as much as 17\% of Indian Facebook posts \citep{Bali14borrowing} and 3.5\% of all tweets \citep{Rijhwani17estimating} are code-switched. This phenomenon is also becoming more pervasive in short text messages, chats, blogs, and the like \citep{Samih16multilingual}. Code-switching however remains understudied in natural language processing (NLP) \citep{Aguilar20english}, and most work to date has focused on token-level language identification (LID) \citep{Samih16multilingual} and on language models for Automatic Speech Recognition \citep{Winata19code}. More tasks are being considered lately, such as Named Entity Recognition \citep{Aguilar18overview}, Part-of-Speech tagging \citep{Ball18partofspeech} or Sentiment Analysis \citep{Patwa20semeval}.% , especially in the area of Machine Translation (MT).

We focus here on another task for CSW texts: Machine Translation (MT). The advent of Neural Machine Translation (NMT) technologies \citep{Bahdanau14neural, Vaswani17attention} has made it possible to design multilingual models capable of translating from multiple source languages into multiple target languages \citep{Firat16multiway,Johnson17googles}, where however both the input and output are monolingual. We study here the ability of such architectures to translate fragments freely mixing a ``matrix'' and an ``embedded'' language into monolingual utterances.

Our main contribution is to show that for the two pairs of languages considered (French-English and Spanish-English): (a) translation of CSW texts is almost as good as the translation of monolingual texts -- a performance that bilingual systems are unable to match; (b) such results can be obtained by training solely with artificial data; (c) CSW translation systems achieve a near deterministic ability to recopy in the output target words found in the input, suggesting that they are endowed with some language identification abilities. Using these models, we are also able to obtain competitive results on the SemEval 2014 Task~5: L2 Writing Assistant, which we see as one potential application area of CSW translation.

\section{Building translation systems for code-switched data\label{sec:cs-systems}}

\subsection{Code-switched data generation\label{subsec:csgen}}

Parallel corpora with natural CSW data are very scarce \citep{Menacer19machine} and, similar to \citet{Song19code}, we generate artificial CSW parallel sentences from regular translation data.

We first compute word alignments between parallel sentences using \texttt{fast\_align}\footnote{\url{https://github.com/clab/fast\_align}} \citep{Dyer13simple}. We then extract so-called \emph{minimal alignment units}  following the approach of \citet{Crego05reordered}: these correspond to small bilingual phrase pairs $(e,f)$ extracted from (symmetrized) word alignments such that all alignment links outgoing from words in $e$ reach a word in $f$, and vice-versa.

For each pair of parallel sentence, we first randomly select the matrix language;\footnote{%
  Note that we abuse here the terms ``matrix'' and ``embedded'' language, as we do not attempt to generate linguistically realistic CSW data matching the constraints of the MLF theory. We use these terms in a much looser sense where the sentence in the ``matrix'' language is the one that receives arbitrary insertions from the ``embedded'' language. This means that our artificial CSW sentences will contain insertions of unconstrained fragments containing both content and function words, which the theory would generally consider ungrammatical.} then the number of replacements $r$ to appear in a derived CSW sentence with an exponential distribution as:
\begin{equation}
P(r=k) = \frac{1}{2^{k+1}} \quad \forall k = 1, \dots, \operatorname{rep}
\label{eq:prob}
\end{equation}
where $\operatorname{rep}$ is a predefined maximum number of replacements. We also make sure that the number of replacements does not exceed half of either the original source or target sentences length, adjusting the actual number of replacements as:
\begin{equation}
n = \min (\frac{S}{2}, \frac{T}{2}, r)
\end{equation}
where
% $r$ is  a random number sampled using Equation~\eqref{eq:prob} and
$S$ and $T$ are respectively the length of the source and target sentences. We finally choose uniformly at random $r$ alignment units and replace these fragments in the matrix language by their counterpart in the embedded language. Figure~\ref{fig:example} displays examples of generated CSW sentences.

\begin{figure*}[ht]
  \center
  \scalebox{0.85}{
  \begin{tabular}{c|l}
  \hline
  Matrix & In Oregon , planners are experimenting with giving drivers different choices . \\
  \hline
  $r=1$    & \textbf{Dans} Oregon , planners are experimenting with giving drivers different choices . \\
  $r=2$    & \textbf{Dans} Oregon , \textbf{les planificateurs} are experimenting with giving drivers different choices . \\
  $r=3$    & \textbf{Dans} Oregon , \textbf{les planificateurs} are experimenting \textbf{en offrant aux} drivers different choices . \\
  \hline
  Embedded & Dans l'Orégon, les planificateurs tentent l'expérience en offrant aux automobilistes différents choix.\\
  \hline
  \end{tabular}
  }
  \caption{Examples of generated CSW sentences when taking English as the matrix language and varying the number $r$ of replacements of embedded French segments (in boldface).\label{fig:example}}
\end{figure*}

\subsection{Machine translation for CSW data \label{sssec:csdata}}
\subsubsection{Data preparation}
We use WMT data for CSW data generation and for training MT systems. We discard sentences which do not possess the correct language by using the \texttt{fasttext} LID model\footnote{\url{https://dl.fbaipublicfiles.com/fasttext/supervised-models/lid.176.bin}} \citep{Bojanowski17enriching}. We use Moses tools \citep{Koehn07moses} to normalize punctuations, remove non-printing characters and discard sentence pairs with a source / target ratio higher than~1.5, with a maximum sentence length of~250. We tokenize all WMT data using Moses tokenizer.\footnote{\url{https://github.com/moses-smt/mosesdecoder}} Our procedure for artificial CSW data generation uses WMT13 En-Es parallel data with 14.5M sentences.
% When training LID models with LinCE data, we exclude the UN data, whose content stronger differs from the tweets in the LinCE corpus. The remaining parallel data contains 3M sentences.
For En-Fr, we use all WMT14 parallel data, for a grand total of 33.9M sentences. Our development sets are respectively \texttt{newstest2011} and \texttt{newstest2012} for En-Es, and  \texttt{newstest2012} and \texttt{newstest2013} as development sets for En-Fr; the corresponding test sets are \texttt{newstest2013} (En-Es) and \texttt{newstest2014} (En-Fr).

% We use the same data generation mechanism to generate CS dev sets from the original dev sets, CS-newstest2013 for En-Es and CS-newstest2014 for En-Fr.

\subsubsection{Machine Translation systems}

We use the \texttt{fairseq}\footnote{\url{https://github.com/pytorch/fairseq}} \citep{Ott19fairseq} implementation of Transformer base \citep{Vaswani17attention} for our models with a hidden size of $512$ and a feedforward size of $2048$. We optimize with Adam, set up with an initial learning rate of $0.0007$ and an inverse square root weight decay schedule, as well as $4000$ warmup steps. All models were trained with mixed precision and a batch size of $8192$ tokens for 300k iterations on $4$ V100 GPUs. 
For each language pair, we use a shared source-target inventory built with Byte Pair Encoding (BPE) of 32K merge operations, using  the implementation published by \citet{Sennrich16BPE}.\footnote{\url{https://github.com/rsennrich/subword-nmt}.} 
%For each language pair, we use a shared source-target inventory of 32K translation units, built with Byte Pair Encoding (BPE), using  the implementation described in \citep{Sennrich16BPE}.\footnote{\url{https://github.com/rsennrich/subword-nmt}.} 
Note that we do not share the embedding matrices. Our experiments with sharing the decoder's input and output embeddings or sharing all encoder+decoder embeddings did not yield further gains.

We compare three settings for \texttt{Code-Switch} models:
\begin{itemize}
\item the \texttt{base-csw} setting, where we train two separate systems, one translating CSW into English, and the other translating CSW into Spanish or French.
\item the \texttt{multi-csw} setting, where we train one model able to generate either pure matrix or embedded language in the output. To this end, similar to a multilingual NMT model \citep{Johnson17googles}, we add a tag at the beginning of each CSW sentence to specify the desired target language. Taking En-Fr as an example, we add a \texttt{<EN>} tag for CSW-En and a \texttt{<FR>} tag for CSW-Fr.  We use the combination of CSW-En and CSW-Fr data for training, which implies that each source side (CSW sentence) is duplicated in the training data, once for each possible output.
\item the \texttt{joint-csw} setting, which extends \texttt{multi-csw} by using one encoder and two separate decoders and training the two output languages simultaneously \emph{with a combined loss function}: for each training (CSW) instance, the loss function sums the two prediction terms for the embedded and the matrix language. The training data remains the same.
\end{itemize}
Note that all our \texttt{Code-Switch} systems also have the ability to translate monolingual source data, in either direction.

For comparison purposes, we also use our parallel data to train two baselines: (a) regular NMT systems for the considered language pairs (\texttt{base}), similar to \texttt{base-csw}; (b) \emph{bilingual NMT systems}, capable of translating from and into both two languages (\texttt{bilingual}). The selection of the desired target language relies on the same tagging mechanism as \texttt{multi-csw}, which means that both types of models see exactly the same examples.
% We train 2 such bilingual models with Transformer base \texttt{bilingual} and Transformer big \texttt{bilingual-big}.
All resulting baseline Transformer models have the exact same hyperparameters and use the same training scheme as \texttt{Code-Switch}.\fyDone{distinguish Matrix/Embedded and Source/Target, also Code-Switch/multi-csw}
% , while \texttt{bilingual-big} has a larger hidden representation of dimension $1024$ and a feedforward size equals to $4096$.
Performance is computed with SacreBLEU \citep{Post18sacrebleu} and METEOR \citep{Denkowski14meteor}.

\section{Machine translation experiments\label{sec:results}}

\subsection{Results\label{ssec:trans}}

\fyDone{Sharing parameters, description of systems}
We run tests using artificial CSW datasets, as mentioned in Section~\ref{sssec:csdata}, as well as on the original test sets, in order to evaluate our models' ability to translate both CSW and monolingual sentences. Results are in Table~\ref{tab:resMT} where we also separately report scores for the `Matrix' and `Embedded' part of the test sets. As is obvious on the \texttt{copy} line, the `Embedded' part contains mostly source language, and corresponds to an actual translation task whereas the `Matrix' part mostly contains target words on the source side, and is much easier to translate.

On  the left part of this table, we see that the baseline systems, either with two (\texttt{base}) or one single (\texttt{bilingual}) model(s), do better on monolingual test sets than their counterparts trained on CSW data (respectively \texttt{base-csw} and \texttt{multi-csw}). For both language pairs, the observed differences are in the range of 1-1.5 BLEU points. Conversely, when translating CSW sentences, \texttt{*-csw} models perform significantly better than the corresponding baselines models, which have never seen CSW in the source.

% This happens when the number of target words in the source is small, in which case the system falls back to its usual ``translation mode'' (rather than employing the ``copy mode'' prescribed by the target language token).
Moreover, we note the marked differences between BLEU scores obtained by these models when the matrix language for the CSW source is the target and when the embedded language is the target. In the former case, translation is near perfect; in the latter case they nonetheless use the little information available to improve over the monolingual scores (about 1-1.5 BLEU points), nearly matching the performance of the baseline systems. This is illustrated for Fr-En, for which \texttt{joint-csw} improved from 33.7 to 35.0; in the same condition, the \texttt{bilingual} system only improves by 0.1 point.

Among the three \texttt{Code-Switch} models, \texttt{multi-csw} is the weakest, while the other two achieve comparable performance. Interestingly, with joint training (\texttt{joint-csw}), we can recover with one single system the performance of the two separate systems used in the \texttt{base-csw} condition. On the monolingual tests, this system also matches the performance of the multilingual baseline (\texttt{bilingual}), which makes it overall our best contender of the lot.

\begin{table*}[!ht]
\center
\scalebox{0.9}{
%\hspace{-1.5cm}
%\begin{tabular}{|l|c|c|c|c|c|c|c|c|}
%\begin{tabular}{lcccccccc}
%\hline
%Testset & \multicolumn{4}{c}{newstest2013} & \multicolumn{4}{c}{cs-newstest2013} \\
%\hline
%Direction & \multicolumn{2}{c}{En-Es} & \multicolumn{2}{c}{Es-En} & \multicolumn{2}{c}{CS-Es} & \multicolumn{2}{c}{CS-En} \\
%\hline
%Metrics & BLEU & METEOR & BLEU & METEOR & BLEU & METEOR & BLEU & METEOR \\
%\hline
%cs-newstest2013 &    -       &     -      &     -     &    -           &  50.3   &       57.8    &  46.8   &    31.7 \\
%\texttt{bilingual} &  31.9 &      57.3    &  32.6  &    35.9      &   23.3  &     42.0     &  44.2  &    37.5 \\
%\texttt{bilingual-big} & 34.2 &  59.1  &  34.3  &    36.9      &   50.6  &     69.3     &  54.6   &   43.9 \\
%\texttt{Code-Switch} & 31.1  & 56.7  &  31.5   &   35.4      &   66.5   &    79.5     &  64.7   &    48.6 \\
%\hline

\begin{tabular}{l|cccccc|cccccc}
\hline
Testset & \multicolumn{6}{c|}{newstest2013} & \multicolumn{6}{c}{csw-newstest2013} \\
\hline
Direction & \multicolumn{3}{c}{En-Es} & \multicolumn{3}{c|}{Es-En} & \multicolumn{3}{c}{CSW-Es} & \multicolumn{3}{c}{CSW-En} \\
\hline
Metrics & \multicolumn{2}{c}{B} & M & \multicolumn{2}{c}{B} & M & \multicolumn{2}{c}{B} & M & \multicolumn{2}{c}{B} & M \\
\hline
\texttt{copy} & \multicolumn{2}{c}{-} & - & \multicolumn{2}{c}{-} & -  &   \multicolumn{2}{c}{50.3} &  57.8 &  \multicolumn{2}{c}{46.8}  &   31.7 \\[-3pt]
                                 &          &               &     &             &                   &     & \small{2.9} & \small{93.5} &  & \small{3.0} & \small{93.3} &     \\
\hline
\texttt{base} & \multicolumn{2}{c}{\textbf{33.2}} & \textbf{58.3} &  \multicolumn{2}{c}{\textbf{33.8}} & \textbf{36.4} & \multicolumn{2}{c}{38.9} & 59.1 & \multicolumn{2}{c}{57.3} & 44.4 \\[-3pt]
                                 & \small{33.1} & \small{-} &     & \small{34.0} & \small{-} &     & \small{32.5} & \small{43.4} &  & \small{34.6} & \small{78.7} &     \\
\texttt{bilingual} & \multicolumn{2}{c}{31.9} & 57.3 &  \multicolumn{2}{c}{32.6} & 35.9 & \multicolumn{2}{c}{23.3} & 42.0 & \multicolumn{2}{c}{44.2} & 37.5 \\[-3pt]
                                 & \small{31.9} & \small{-} &     & \small{32.9} & \small{-} &     & \small{32.3} & \small{14.5} &  & \small{33.3} & \small{54.5} &     \\
% \texttt{bilingual-big} & \multicolumn{2}{c}{34.2} & 59.1 & \multicolumn{2}{c}{34.3} & 36.9 & \multicolumn{2}{c}{50.6} & 69.3 & \multicolumn{2}{c}{54.6} & 43.9 \\[-3pt]
%                                  & \small{34.0} & \small{34.3} &     & \small{34.6} & \small{34.1} &     & \small{34.9} & \small{73.8} &  & \small{34.4} & \small{65.2} &     \\
\hline
\texttt{base-csw} & \multicolumn{2}{c}{32.0} & 57.4 & \multicolumn{2}{c}{32.7} & 36.0 & \multicolumn{2}{c}{66.8} & \textbf{79.8} & \multicolumn{2}{c}{\textbf{66.5}} & \textbf{49.4} \\[-3pt]
                                 & \small{31.8} & \small{-} &     & \small{33.0} & \small{-} &     & \small{33.1} & \small{97.1} &  & \small{34.5} & \small{97.5} &     \\
\texttt{multi-csw} & \multicolumn{2}{c}{31.1} & 56.7 & \multicolumn{2}{c}{31.5} & 35.4 & \multicolumn{2}{c}{66.5} & 79.5 & \multicolumn{2}{c}{64.7} & 48.6 \\[-3pt]
                                 & \small{30.9} & \small{-} &     & \small{31.9} & \small{-} &     & \small{32.2} & \small{97.2} &  & \small{33.1} & \small{95.1} &     \\
\texttt{joint-csw} & \multicolumn{2}{c}{31.9} & 57.3 & \multicolumn{2}{c}{32.6} & 36.0 & \multicolumn{2}{c}{\textbf{66.9}} & 79.7 & \multicolumn{2}{c}{66.4} & \textbf{49.4} \\[-3pt]
                                 & \small{32.0} & \small{-} &     & \small{32.8} & \small{-} &     & \small{33.2} & \small{97.2} &  & \small{34.2} & \small{97.5} &     \\
\hline
%\end{tabular}
%}
%\scalebox{0.9}{
%\begin{tabular}{lcccccccccccc}
\hline
Testset & \multicolumn{6}{c|}{newstest2014} & \multicolumn{6}{c}{csw-newstest2014} \\
\hline
Direction & \multicolumn{3}{c}{En-Fr} & \multicolumn{3}{c|}{Fr-En} & \multicolumn{3}{c}{CSW-Fr} & \multicolumn{3}{c}{CSW-En} \\
\hline
Metrics &  \multicolumn{2}{c}{B} & M & \multicolumn{2}{c}{B} & M & \multicolumn{2}{c}{B} & M & \multicolumn{2}{c}{B} & M \\
\hline
\texttt{copy} & \multicolumn{2}{c}{-} & -   & \multicolumn{2}{c}{ -} & -   & \multicolumn{2}{c}{50.0} & 55.7 & \multicolumn{2}{c}{46.5}   &  33.1    \\[-3pt]
                                &            &                &    &               &                   &  & \small{2.9} & \small{93.8} &    & \small{2.9} & \small{93.4} &      \\ 
\hline
\texttt{base} & \multicolumn{2}{c}{\textbf{37.9}} & \textbf{60.9} & \multicolumn{2}{c}{\textbf{35.4}} & \textbf{37.9} & \multicolumn{2}{c}{45.1} & 64.4 & \multicolumn{2}{c}{61.3} & 47.3 \\[-3pt]
                                & \small{37.7} & \small{-} &    & \small{35.3} & \small{-} &  & \small{37.8} & \small{52.0} &    & \small{36.0} & \small{84.6} &      \\ 

\texttt{bilingual} & \multicolumn{2}{c}{36.3} & 59.6 & \multicolumn{2}{c}{34.5} & 37.6 & \multicolumn{2}{c}{54.8} & 71.3 & \multicolumn{2}{c}{56.5} & 45.8 \\[-3pt]
                               & \small{36.4} & \small{-} &    & \small{34.6} & \small{-} &  & \small{36.8} & \small{71.7} &    & \small{34.7} & \small{76.6} &      \\ 
% \texttt{bilingual-big} & \multicolumn{2}{c}{39.4} & 62.4 & \multicolumn{2}{c}{37.2} & 38.8 & \multicolumn{2}{c}{41.1} & 57.0 & \multicolumn{2}{c}{55.8} & 44.2 \\[-3pt]
%                                 & \small{39.3} & \small{39.4} &    & \small{37.2} & \small{37.2} &  & \small{40.0} & \small{41.7} &    & \small{37.3} & \small{72.6} &      \\   
\hline
\texttt{base-csw} & \multicolumn{2}{c}{36.7} & 59.9 & \multicolumn{2}{c}{34.3} & 37.5 & \multicolumn{2}{c}{\textbf{67.5}} & \textbf{79.9} & \multicolumn{2}{c}{\textbf{67.9}} & \textbf{50.5} \\[-3pt]
                                & \small{36.7} & \small{-} &    & \small{34.2} & \small{-} &  & \small{37.8} & \small{95.2} &    & \small{35.6} & \small{97.4} &      \\       

 \texttt{multi-csw} & \multicolumn{2}{c}{35.2} & 58.7 & \multicolumn{2}{c}{32.9} & 36.8 & \multicolumn{2}{c}{66.7} & 79.5 & \multicolumn{2}{c}{65.8} & 49.4 \\[-3pt]
                               & \small{35.3} & \small{-} &    & \small{32.6} & \small{-} &  & \small{36.3} & \small{95.1} &    & \small{33.7} & \small{94.6} &      \\
  
\texttt{joint-csw} & \multicolumn{2}{c}{36.2} & 59.5 & \multicolumn{2}{c}{34.0} & 37.3 & \multicolumn{2}{c}{67.4} & 79.8 & \multicolumn{2}{c}{67.7} & 50.3 \\[-3pt]
                                & \small{36.2} & \small{-} &    & \small{33.7} & \small{-} &  & \small{37.3} & \small{95.4} &    & \small{35.0} & \small{97.4} &      \\       

  \hline
\end{tabular}
}

\caption{\label{tab:resMT} Translating monolingual newstest data and artificial \texttt{csw-newstest} data for two language pairs where performance is measured via the BLEU (B) and METEOR (M) scores. We also report a trivial baseline that just recopies the source text. Small numbers contain BLEU scores computed separately when the target language is the embedded language (left) and the matrix language (right). For the monolingual tests (left part), these correspond to scores computed on the same sentences that are also included in the CSW tests.}
\end{table*}
\fyDone{Comment - error analysis - why don't we improve?}
%\begin{table*}[!ht]
%\center
% \scalebox{0.9}{
%\hspace{-1.5cm}
%\begin{tabular}{lcccccccc}
%\hline
%Testset & \multicolumn{4}{c}{newstest2014} & \multicolumn{4}{c}{cs-newstest2014} \\
%\hline
%Direction & \multicolumn{2}{c}{En-Fr} & \multicolumn{2}{c}{Fr-En} & \multicolumn{2}{c}{CS-Fr} & \multicolumn{2}{c}{CS-En} \\
%\hline
%Metrics & BLEU & METEOR & BLEU & METEOR & BLEU & METEOR & BLEU & METEOR \\
%\hline
%cs-newstest2014 &     -     &    -        &   -       &     -           &  50.0   &    55.7      &  46.5   &  33.1    \\
%\texttt{bilingual} & 36.3 &       59.6    & 34.5   &     37.6      & 54.8   &    71.3      &  56.5  &    45.8 \\
%\texttt{bilingual-big} & 39.4 & 62.4   & 37.2   &     38.8     & 41.1   &   57.0       &  55.8  &   44.2 \\
%\texttt{Code-Switch} & 35.2 &  58.7  & 32.9   &     36.8       &  66.7   &    79.5      &   65.8  &    49.4 \\
%\hline

%\caption{\label{tab:resEnFr} Translating original (monolingual) newstest data and artificial (code-switched) cs-newstest data for the language pair En-Fr. We also report the BLEU (B) and METEOR (M) scores of the \emph{source} cs-newstest. Small numbers contain BLEU scores computed separately when the matrix language is in the source (left) and in the target (right).} \fyDone{Maybe somewhere the baseline that does nothing would help}\fyDone{Any reason why big is so much worse than small?}\fyDone{Comment on the difference between matrix / embed language}\fyDone{Check numbers for bilingual systems}
% \end{table*}

\subsection{Analysis\label{sec:analysis}}

\subsubsection{Code-Switching effect\label{ssec:csEffect}}

\begin{figure*}[ht]
  \center
  \begin{subfigure}[t]{0.48\textwidth}
    \includegraphics[width=\textwidth]{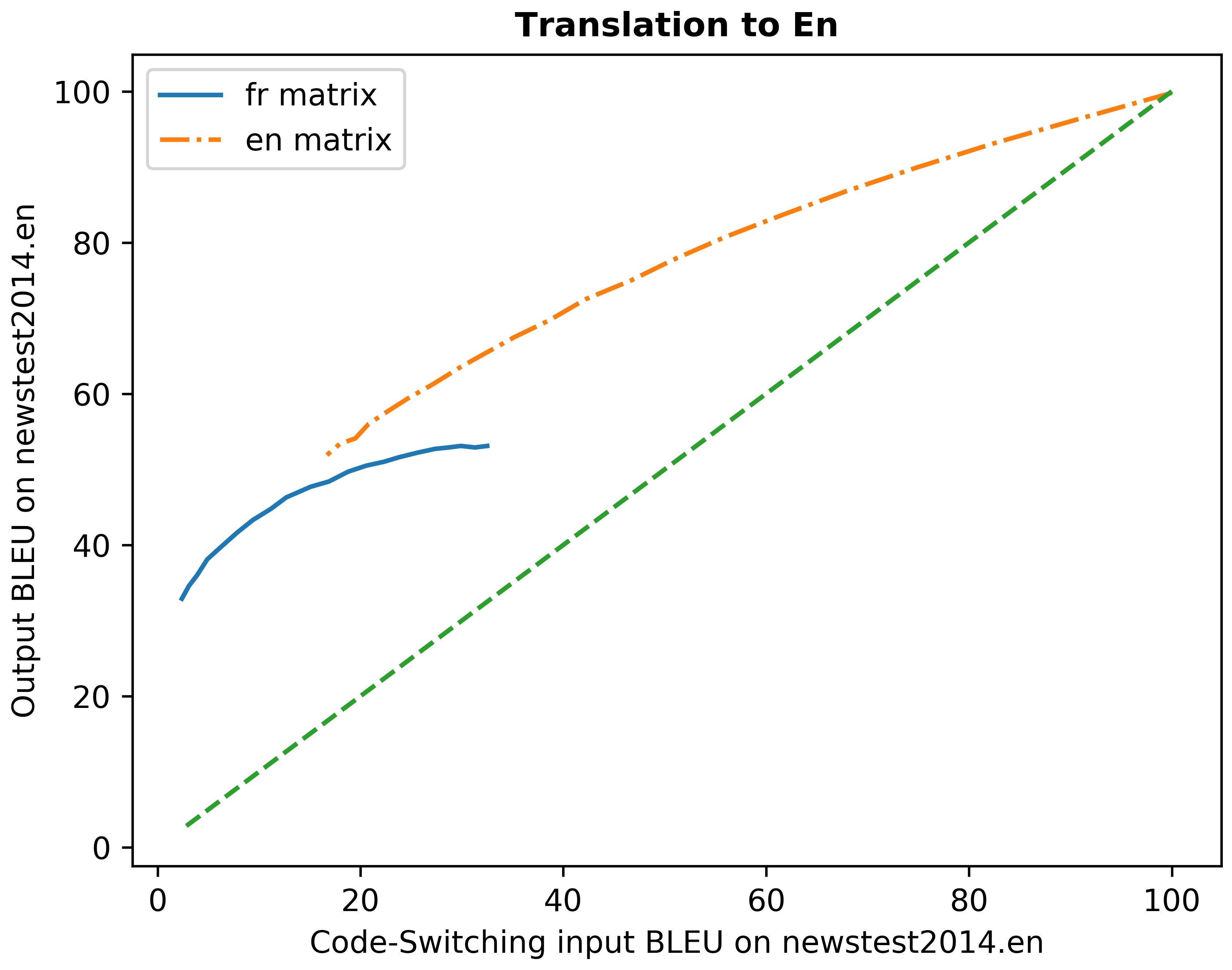}
    \caption{\label{fig:bleuEnFrEn}}
  \end{subfigure}
  \begin{subfigure}[t]{0.48\textwidth}
    \includegraphics[width=\textwidth]{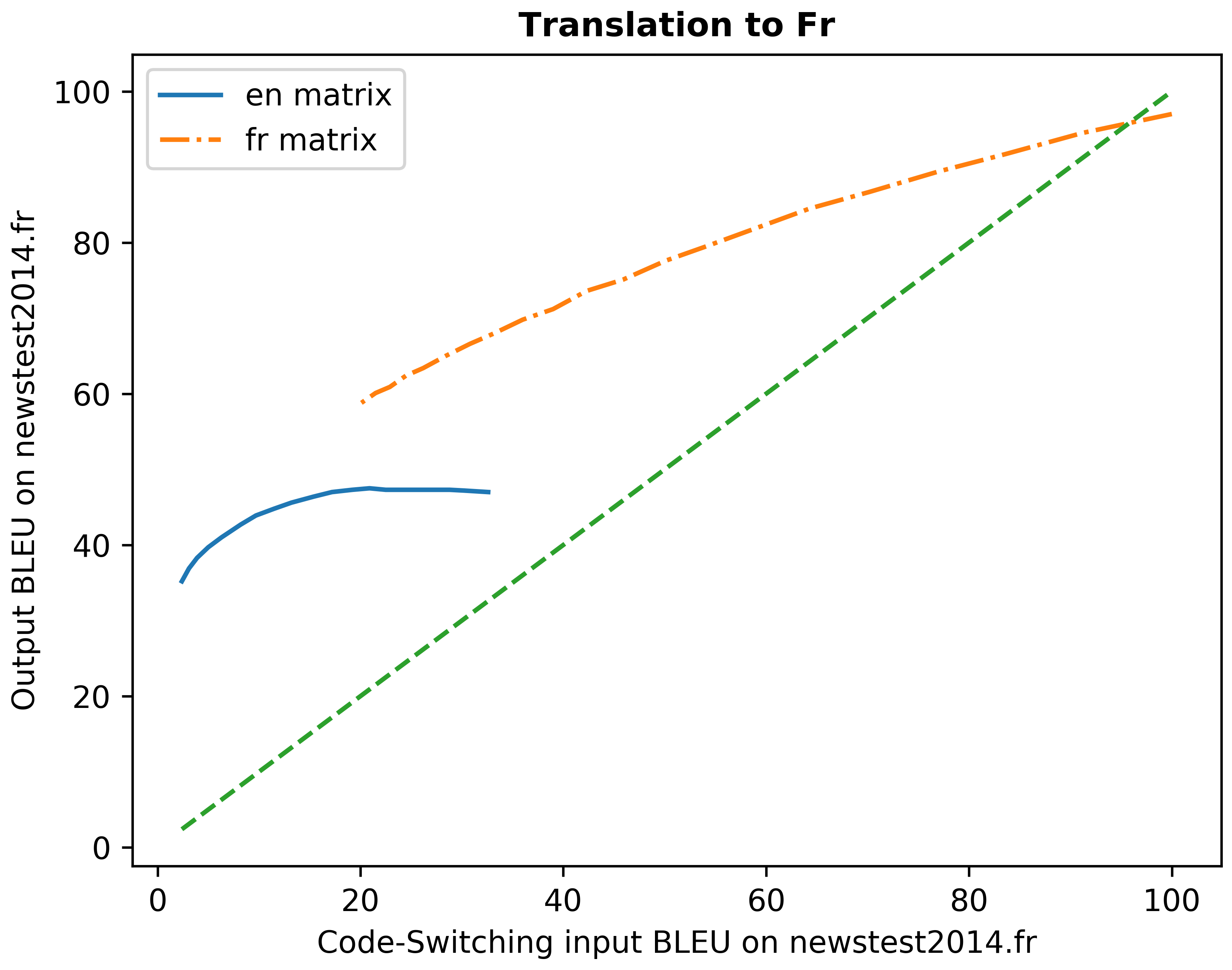}
    \caption{\label{fig:bleuEnFrFr}}
  \end{subfigure}
  \caption{Evolution of the BLEU score of source CSW data and their target translation for En-Fr. (a) Direction CSW-En. The solid curve takes Fr as the matrix language, where we progressively inject more En segments;  for the dash dot curve, En is the matrix language, with a growing number of Fr segments. (b) Direction CSW-Fr. Note that the target BLEU is always much higher than the source BLEU, with about a 20 points difference. The gap between the dash dot and solid curves is due to the basic sentence structure of the matrix language (see Section~\ref{ssec:csEffect}). As dash dot curves represent insertion in the \emph{reference target} sentence, the corresponding BLEU score is always higher than the solid curve and actually reaches 100 (in the absence of any embedded language).\label{fig:bleuEnFr}}\fyDone{Explain what we see left and right}\fyDone{Change legend (over) in Figures}\fyDone{Change colours to patterns}%the limit on the replacement ratio 
\end{figure*}

In order to better study the effect of mixing languages, we modify the synthetic data generation method to keep one language as the matrix language, in which segments are incrementally replaced by translations of the embedded language.  We relax the constraint on the maximum number of replacements and generate new test sets with an increasing number of replacements, ranging from 1 to 20, resulting in 20\footnote{For sentences that could not accommodate 20 replacements, we performed as many replacements as possible.} versions of the CSW test sets (in each direction).
In Figure~\ref{fig:bleuEnFr},
   %    and \ref{fig:bleuEnEs},
we plot the BLEU scores of both source CSW sentences and their translations for En-Fr language pair, using each language as the matrix language, to visualize the impact of progressively introducing more target fragments into the source.

% \begin{figure*}[ht]
%   \center
%   \begin{subfigure}[t]{0.45\textwidth}
%     \includegraphics[width=\textwidth]{fig/BLEU.en-es.en}
%     \caption{\label{fig:bleuEnEsEn}}
%   \end{subfigure}
%   \begin{subfigure}[t]{0.45\textwidth}
%     \includegraphics[width=\textwidth]{fig/BLEU.en-es.es}
%     \caption{\label{fig:bleuEnEsEs}}
%   \end{subfigure}
%   \caption{Evolution of the BLEU score of source CS data and their target translation for EN-ES. (a) refers to direction CS-En with Es as the matrix language for solid curve and En for dashdot curve. (b) refers to direction CS-Es with En as the matrix language for solid curve and Es for dashdot curve.\label{fig:bleuEnEs}}
% \end{figure*}

The same behavior is observed for both language pairs and directions: on average, inserting random target fragments boosts the translation performance, with a larger payoff for the first few target segments. There exists an important gap for the output BLEU scores when CSW source sentences with different matrix languages reach the same (input) BLEU scores. Even though we generate a large number of replacements, the basic grammar structure of the matrix language is still maintained. Therefore, taking the target language as matrix gives the model a pre-translated sentence structure that is much easier to reproduce.

\subsubsection{Implicit LID in translation}

A second question concerns the ability of the translation system to identify target fragments in the source and to copy them in the target, even though these fragments are indistinguishable from genuine source segments. We use labels computed during the CSW generation procedure to sort out pre-translated (target) segments from actual source segments to be translated. For instance, when translating into French, only tokens with a label \texttt{eng}, denoting English, are expected to be translated. All other tokens correspond to French words are expected to be copied. As reported in Table~\ref{tab:resCopy}, our translation models are able to copy almost all pre-translated tokens for both language pairs and directions.

Refining the analysis, we also study whether the relative order of target words changes, or is preserved, during the translation. Table~\ref{tab:resOrder} reports the percentage of exact and switched-order copies. We observe again large differences with respect to the position of the matrix language. When the matrix language is the target language, the model always preserves the observed token order since it indicates a correct sentence structure for the hypothesis. When translating into the embedded language, we observe a larger number of word order changes: in this case, inserted target segments may not appear in their correct order in the CSW sentence, an issue that the model tries to fix. An example of this is in Figure~\ref{fig:noise}, where we observe a swap between the input (``différent choix'') and output (``choix différent'') word orders.
% For instance, ``French government'' is replaced by 2 segments ``fran\c{c}ais'' and ``governement'' in the source sentence and the model corrects the order by predicting ``governement fran\c cais''.
%This suggests that the MT system implicitly performs some language identification during translation.
\done\Todo{Does the system generate CS data ? which proportion of translation in English is not English ?}
\done\Todo{Does the system copy what is pre-translated ? \fyTodo{Do we see changes in the pre-translated segments?}}
\done\Todo{Are there cases where small amount of pre-translation changes a lot in BLEU ?}

\begin{table}[!ht]
\center
\scalebox{0.8}{
\begin{tabular}{lcccc}
\hline
%	& Nb inexact copy (in 3000) & Sum of uncopied tokens &  Avg Copy rate (\%) per sentence \\
%	& EN & ES \\
Testset & \multicolumn{2}{c}{csw-newstest2014} & \multicolumn{2}{c}{csw-newstest2013}  \\
\hline
Direction & CSW-En & CSW-Fr & CSW-En & CSW-Es \\
\hline
%EN & 43 & 48 & 91.31  \\
%\hline
%ES & 53 & 69 & 94.23 \\
%\hline
%Nb inexact copy (in 3000) & 43 & 53 \\
%Sum of uncopied tokens & 48 & 69 \\
%Avg Copy rate (\%) per sentence & 91.31 & 94.23 \\
to copy & 42148 & 47337 & 37653 & 41053 \\
copied  & 41567 & 46229 & 37421 & 40638 \\
copy rate (\%)   &  98.6   &   97.7  &   99.4  & 99.0 \\
CSW rate (\%)     &  0.13   &  0.30   &   0.16  &  0.23 \\
\hline
\end{tabular}
}
\caption{\label{tab:resCopy} Analyzing the recopy of tokens on \texttt{csw-newstest2014} for En-Fr and \texttt{csw-newstest2013} for En-Es. We report the number of (pre-translated) tokens that should be copied, and the corresponding ratios.}
\end{table}

\begin{table}[!ht]
\center
\scalebox{0.8}{
\begin{tabular}{lcc|cc}
  & \multicolumn{2}{c}{En-Fr} & \multicolumn{2}{c}{En-Es} \\
    \cline{2-5}
Direction & Copy & Copy+Swap & Copy & Copy+Swap \\ %& Tokens changed \\
\hline
CSW-En                & 87.1 & 4.5    & 90.7 & 5.2     \\ %& 8.4 \\
$\quad$ Mat En & 97.6 & 0.1    & 98.2 & 0.2     \\ %& 2.2 \\
$\quad$ Mat For  & 61.4 & 15.2  & 72.6 & 17.3    \\ %& 23.6 \\
CSW-For                & 77.4 & 5.5   & 88.5 & 3.7   \\ %& 17.1 \\
$\quad$ Mat For  & 84.9 & 0.1   & 97.1 & 0.2   \\ %& 15.0 \\
$\quad$ Mat En & 59.4 & 18.5 & 65. 6 & 13.2  \\ %& 22.1 \\
% \hline
 % CS-En                & 90.7 & 5.2   & 4.1  \\
 % $\quad$ Mat En & 98.2 & 0.2   & 1.6   \\
 % $\quad$ Mat Es & 72.6 & 17.3  & 10.1  \\
 % CS-Es                & 88.5 & 3.7   & 7.8    \\
 % $\quad$ Mat Es & 97.1 & 0.2   & 2.7   \\
 % $\quad$ Mat En & 65.6 & 13.2 & 21.2  \\
\hline
\end{tabular}
}
\caption{\label{tab:resOrder} Percentage of sentences for which all target words have been exactly copied without and with order changes, for \texttt{csw-newstest2014} (En-Fr) and \texttt{csw-newstest2013} (En-Es). We separately report numbers for the case where the foreign language (French or Spanish) is the embedded (Mat En) or matrix (Mat For) language.}
\end{table}

Conversely, it is also interesting to look at the proportion of mixed language generated on the target side. Recall that in our training, the source is mixed-language, while the target is always monolingual. We use an in-house token-level language identification (\texttt{LID}) model to identify the language of output tokens and to detect the CSW rate on the target side. As indicated in Table~\ref{tab:resCopy}, our models generate almost pure monolingual translations, with a very low rate of CSW text.
% Note that the detected CS rate is much lower than the \texttt{LID} model's level of accuracy, as shown in Table \ref{tab:lid}. \fyDone{Results ?}
CSW-translation models thus seem to perform some language identification, as they almost perfectly sort out target language tokens (which are almost always copied) from the source language tokens (which are always translated).
% and turn out to be even more precise than a specifically trained \texttt{LID} model (compare Table~\ref{tab:resCopy} to Table~\ref{tab:lid}).

\begin{figure*}[ht]
  \center
  \scalebox{0.8}{
  \begin{tabular}{c|l}
  \hline
  En & In Oregon , planners are experimenting with giving drivers different choices. \\
  Fr & Dans l'Orégon, les planificateurs tentent l'expérience en offrant aux automobilistes différents choix.\\
  \hline
  CSW   & In l'Oregon , planners \textbf{tentent l' expérience} with giving \textbf{automobilistes différents choix}. \\
  Hyp & \textit{Dans l'Orégon , les planificateurs tentent l'expérience de donner aux automobilistes différents choix.} \\
  Noisy CSW & In l' Oregon , planners \textbf{tenter} l'expérience with giving \textbf{automobilist} \textbf{différent} choix.  \\
  Hyp & \textit{Dans l'Orégon , les planificateurs \textbf{doivent tenter} l'expérience de donner \textbf{à l' automobiliste} \textbf{un choix différent}. }\\
  \hline
  \end{tabular}
  }
  \caption{A noisy Code-Switched sentence with French as both the matrix and target language.\label{fig:noise}}
\end{figure*}

A last issue concerns morphological errors: when inserting foreign words into a matrix source, one cannot expect to always also introduce the right inflection marks, some of which can only be determined once the target context is known. Another interesting phenomenon, that we do not simulate here, is when the embedded (target) lemma is adapted bears a morphological mark that only exist in the matrix language, which means that two linguistic systems are mixed within the same word, thereby posing more extreme difficulties for MT \citep{Manandise11morphology}. 

To illustrate the ability to correct grammar errors in input fragments, we manually noise a CSW sentence and display its translation in Figure~\ref{fig:noise}. Where the input just contains the  lemma of the French word ``tenter''  ({\sl to try}), the model inserts a modal ``doivent'' to fix the context. Another illustration is for the adjective ``différent'' which is moved into post-nominal position, and for which an article (``un'') is inserted. This indicates that the model not only copies what already exists but also tends to adjust translations whenever necessary.

\fyFuture{This is interesting - may be we could do more systematic texts}

\section{Computing translations in context\label{sec:semeval}}

In this section, we evaluate CSW translation for the SemEval~2014 Task~5: L2 Writing Assistant \citep{vanGompel14semeval}, which can be handled as an MT task from mixed data.\fyDone{Explain}

\subsection{Method}
This task consists in translating L1 fragments in an L2 context, where the test set design is such that there is exactly one L1 insert in each utterance. We evaluated on two L1-L2 pairs: English-Spanish and French-English, and list below example test segments provided by the organizers for these pairs of languages (the insert and reference segments are in boldface):
\begin{itemize}
\item Input (L1=English,L2=Spanish): \textit{“Todo ello, \textbf{in accordance} con los principios que siemprehemos apoyado.”} \\Output: \textit{“Todo ello, \textbf{de conformidad} con los principios que siempre hemos apoy-ado.”}
\item Input (L1=French,L2=English): \textit{“I \textbf{rentre à la maison} because I am tired.”} \\Output: \textit{“I \textbf{return home} because I am tired.”}
\end{itemize}

The official metric for the SemEval evaluation is a word-based accuracy of the translations of the L1 fragment, which  %    instead of the most commonly used BLEU score.
means that the L2 context of each sentence is not taken into account in scoring.  Since our systems are full-fledged NMT systems, their output may not contain the reference L2 prefix and suffix. Therefore, two options are explored to compute these scores. The first is to post-process the output \texttt{HYP} and align it with the L2 reference context in \texttt{REF}. This alignment allows us to only score the relevant fragment in \texttt{HYP}. We refer to this option as \texttt{free-dec}.
% which is aligned to the fragment to be evaluated in \texttt{REF}.

The second option is to ensure that the L2 context will be present in the output translation. To this end, we use the \emph{force decoding mode} of \texttt{fairseq}, implementing the methods of \citet{Post18constraint,Hu19improved}. We explored two different ways to express the L2 context as decoding constraints. The first turns every token in the L2 context as a separate constraint (\texttt{token-cst}). Continuing the previous example, ``\textit{I, because, I, am, tired.}"  yield 5~constraints. The second uses the prefix and suffix of the L2 context as two multi-word constraints (\texttt{presuf-cst}). In this case, ``\textit{I}'' and ``\textit{because I am tired.}'' yield just 2 constraints. In both cases, constraints are required to be present in the prescribed order in the output.

\subsection{Results\label{ssec:semevalres}}
Scores are computed with the SemEval evaluation tool,\footnote{\url{https://github.com/proycon/semeval2014task5}} which enables a comparison with other submissions for this task. Results are in Table~\ref{tab:resSemeEnEs} and \ref{tab:resSemeEnFr}. For En-Es, our CSW translator outperforms the best system in the official evaluation \citep{vanGompel14semeval}. Note that this model is not specifically designed nor tuned in any way for the SemEval task. For Fr-En, our system achieves better performance than the forth best participating system, with a clear gap with respect to the top results. In both cases, constraint decoding hurts performance: given that the automatic copy of target segments is already nearly perfect, introducing more constraints during the search has here a clear detrimental effect for this task.

\begin{table}[!ht]
\center
\scalebox{0.8}{
\begin{tabular}{lccc}
\hline
	& Accuracy & Word Accuracy & Recall \\
\hline
UEdin-run2 & 0.755 & 0.827 & 1.0 \\
UEdin-run1 & 0.753 & 0.827 & 1.0 \\
UEdin-run3 & 0.745 & 0.820 & 1.0 \\ 
\hline
% \textbf{Our model} & 0.741 & 0.815 & 0.998 \\
\texttt{multi-csw} & & & \\
\texttt{free-dec} & 0.755 & 0.827 & 1.0 \\
%\hline
%UNAL-run2& 0.733 & 0.809 & 0.994 \\
\texttt{token-cst} & 0.749 & 0.824 & 1.0 \\
\texttt{presuf-cst} & 0.751 & 0.827 & 1.0 \\
\hline
\texttt{joint-csw} & & & \\
\texttt{free-dec} & 0.773 & 0.842 & 1.0 \\
\hline
\end{tabular}
}
\caption{\label{tab:resSemeEnEs} Results of SemEval 2014 Task 5 for En-Es.}
\end{table}

\begin{table}[!ht]
\center
\scalebox{0.8}{
\begin{tabular}{lccc}
\hline
	& Accuracy & Word Accuracy & Recall \\
\hline
UEdin-run1 & 0.733 & 0.824 & 1.0 \\
UEdin-run2 & 0.731 & 0.821 & 1.0 \\
UEdin-run3 & 0.723 & 0.816 & 1.0 \\ 
CNRC-run1 & 0.556 & 0.694 & 1.0 \\
\hline
% Our model & 0.444 & 0.596 & 0.990 \\
\texttt{multi-csw} & & & \\
\texttt{free-dec} & 0.554 & 0.685 & 0.996 \\
\texttt{token-cst} & 0.531 & 0.665 & 0.990 \\
\texttt{presuf-cst} & 0.519 & 0.658 & 0.982 \\
\hline
\texttt{joint-csw} & & & \\
\texttt{free-dec} & 0.626 & 0.744 & 0.994 \\
\hline
\end{tabular}
}
\caption{\label{tab:resSemeEnFr} Results of SemEval 2014 Task 5 for Fr-En.} %Low accuracy mostly come from non-domain adaptation.}
\end{table}

To better study the performance gap between these language pairs, we additionally score the development and test data with BLEU and METEOR. Results in Table~\ref{tab:resSeme} show that for these metrics, we achieve performance that are in that same ballpark for the two language pairs, suggesting that the observed difference in the SemEval metric is likely due to a mismatch between references and system outputs. The official metric is a word accuracy which may exclude acceptable translations by exact token match. \fyDone{Why? What kind of errors?}
\done\Todo{Maybe add examples about errors.?}

\begin{table}[!ht]
\center
\scalebox{0.9}{
\begin{tabular}{lcccc}
%\begin{tabular}{lcc}
  \hline
  & \multicolumn{2}{c}{\texttt{multi-csw}} & \multicolumn{2}{c}{\texttt{joint-csw}} \\
Dataset & B & M & B & M \\
\hline
Fr-En dev & 97.3 & 75.6 & 97.6 & 76.4 \\
Fr-En test & 90.1 & 64.1 & 91.0 & 66.1 \\
En-Es dev & 97.4 & 98.8 & 97.6 & 99.0 \\
En-Es test & 89.9 & 95.3 & 90.4 & 95.5 \\
\hline
\end{tabular}
}
\caption{\label{tab:resSeme} Results of other metrics on SemEval data. METEOR scores for the Fr-En SemEval test are much worse than for En-Es. This is mostly due to the high ``fragmentation penalty'' computed by METEOR for English; the corresponding average $F_{mean}$ is about 0.99, showing that translations are mostly correct.}
\end{table}

\section{Related work \label{sec:related}}

\fyDone{Transition for this section}
Research in the area of NLP for CSW has mostly focused on CSW Language Modeling, especially for Automatic Speech Recognition \citep{Pratapa18language,Garg18code,Gonen19language,Winata19code,Lee20modeling}. Evaluation tasks, benchmarks have also been prepared for LID in user generated CSW content \citep{Zubiaga16tweelid,Molina16overview}, Named Entity Recognition \citep{Aguilar18overview}, Part-of-Speech tagging \citep{Ball18partofspeech,Aguilar20lince,Khanuja20gluecos} and Sentiment Analysis \citep{Patwa20semeval}. CSW was also found useful in foreign language teaching: \citet{Renduchintala19simple,Renduchintala19spelling} showed that replacing words by their counterparts in foreign language helps to learn foreign language vocabulary.

Regarding MT, most past work has focused on using artificial CSW  data to help conventional translation systems. \citet{Huang14improving} used CSW corpus to improve word alignment and statistical MT. \citet{Dinu19training} experienced replacing and concatenating source terminology constraints by the corresponding translation(s) to boost the accuracy of term translations. \citet{Song19code} shared the same idea by replacing phrases with pre-specified translation to perform ``soft" constraint decoding. A different line of research is in \citep{Bulte19fuzzy,Xu20boosting,Pham20priming}, who explore ways to combine a source sentence with similar translations extracted from translation memories. \citet{Yang20csp} also pre-trained translation models by predicting original source segments from generated CSW sentences and claimed better results compared to other pre-training methods \citep{Conneau2019xlm,Song19mass}. Nevertheless, there barely exists work aimed at translating CSW sentences. \citet{Johnson17googles} mentioned using a multilingual NMT system to translate CSW sentence to a third target language by showing only one example. To the best of our knowledge, only one parallel Arabic-English CSW corpus was specifically released for MT applications \citep{Menacer19machine}. This CSW data was extracted from the UN data with Arabic as the matrix language: while translations into English were readily available, the purely Arabic side of the corpus was obtained using Google Translate to fill the missing Arabic bits.

\section{Conclusion and outlook\label{sec:conclusion}}
\fyDone{Write conclusion}
In this study, we present a data augmentation method to generate artificial CSW data. We have shown that artificial data generated could be used to train NMT systems to translate both monolingual and CSW sentences (in one or even two different languages). With joint training of the two languages, we were able to build systems that were as good as a baseline bilingual system on monolingual texts, and much better for CSW texts. Our system does not need any explicit language identification and almost perfectly sorts out source tokens from target tokens in a CSW utterance. Another interesting feature of our system is that it always output monolingual translations. We finally report state-of-the-art results for the SemEval L2 Writing Assistant task for Es-En, while the related results for Fr-En are still somewhat lagging behind the best scores. 

In the future, we would like to generate more realistic CSW data from monolingual sentences using a translation model. We also plan to explore ways to translate CSW texts simultaneously into both languages, so that the two decoding processes can mutually influence one another: in a first step in that direction, we have shown that training with a joint loss was actually beneficial for the translation into the two languages. Another line of research would be to continue experimenting with realistic language data, also containing other phenomena such as morphological binding. Finally, we also intend to study the somewhat more realistic condition where a mixture of languages A and B is translated into language C; we believe that the artificial CSW generation methods developed in our work would also be effective for this task.%, whereby the inflections in one l
% seeking  an whether the translation in different language could help each other in decoding. \fyDone{Future Work ?}

\section{Acknowledgements\label{sec:acknowledgements}}
% Jean Zay
%This work was performed using the HPC resources of the Jean-Zay computing platform from GENCI–IDRIS.
This work was granted access to the HPC resources of IDRIS under the allocation 2021-[AD011011580R1] made by GENCI. The authors wish to thank Josep Crego for his comments of an earlier version of this work. We also would like to thank the anonymous reviewers for their valuable suggestions. The first author is partly funded by Systran and by a grant from Région Ile-de-France. 

\bibliography{cstranslation}
\bibliographystyle{acl_natbib}

%\todos{}

\end{document}